\documentclass{article}

\usepackage{PRIMEarxiv}
\usepackage{amsthm}
\usepackage{amsmath}
\usepackage[utf8]{inputenc} % allow utf-8 input
\usepackage[T1]{fontenc}    % use 8-bit T1 fonts
\usepackage{hyperref}       % hyperlinks
\usepackage{url}            % simple URL typesetting
\usepackage{booktabs}       % professional-quality tables
\usepackage{amsfonts}       % blackboard math symbols
\usepackage{nicefrac}       % compact symbols for 1/2, etc.
\usepackage{microtype}      % microtypography
\usepackage{lipsum}
\usepackage{fancyhdr}       % header
\usepackage{graphicx}       % graphics
\graphicspath{{media/}}     % organize your images and other figures under media/ folder

\usepackage{array}
\usepackage{multirow}
\usepackage{multicol}
\usepackage[normalem]{ulem}
\useunder{\uline}{\ul}{}
\title{SaSDim:Self-Adaptive Noise Scaling Diffusion Model \\ for Spatial Time Series Imputation
}
\author{
  Shunyang Zhang, \\
  Central South University \\
  ChangSha
   \And
  Senzhang Wang \\
  Central South University \\
  ChangSha
   \And
  Xianzhen Tan \\
  Central South University \\
  ChangSha
   \And
  Ruochen Liu \\
  Central South University \\
  ChangSha
   \And
  Jian Zhang \\
  Central South University \\
  ChangSha
   \And
  Jianxin Wang \\
  Central South University \\
  ChangSha
  }
\usepackage{amsthm}

\begin{document}
\maketitle

\begin{abstract}

Spatial time series imputation is critically important to many real applications such as intelligent transportation and air quality monitoring. Although recent transformer and diffusion model based approaches have achieved significant performance gains compared with conventional statistic based methods, spatial time series imputation still remains as a challenging issue due to the complex spatio-temporal dependencies and the noise uncertainty of the spatial time series data. Especially, recent diffusion process based models may introduce random noise to the imputations, and thus cause negative impact on the model performance. To this end, we propose a self-adaptive noise scaling diffusion model named SaSDim to more effectively perform spatial time series imputation. Specially, we propose a new loss function that can scale the noise to the similar intensity, and propose the across spatial-temporal global convolution module to more effectively capture the dynamic spatial-temporal dependencies. Extensive experiments conducted on three real world datasets verify the effectiveness of SaSDim by comparison with current state-of-the-art baselines.

\end{abstract}

% keywords can be removed

\section{Introduction}
Spatial time series is a type of data that describes the dynamic relationships within and between locations or regions distributed across space, such as the traffic flow time series data collected from a set of road sensors deployed in a road network. To capture the observations of spatial time series, generally a significant number of sensors, such as cameras and traffic sensors, need to be deployed throughout the space under study. However, complete spatial time series is usually unavailable due to various factors, including sensor malfunctions, unstable communication signals, non-uniform sensor distribution, and the stochastic and dynamic nature of data, leading to data anomalies or missing which is also regarded as data sparsity \cite{wang2016estimating}. Therefore, how to impute the incomplete and sparse spatial time series data is a primary challenging issue that needs to be well addressed before further analysis on the data.

Traditionally, statistical methods, such as ARIAM and HR, are adopted for modeling time series. To further capture the spatial correlations of the time series data collected from different locations matrix completion and tensor decomposition approaches are widely adopted and have shown promising performance. However, due to the linear property of statistical models, capturing complex spatiotemporal relationships still remains challenging. 
\begin{figure}[!t]
\centering
\includegraphics[width=0.7\linewidth]{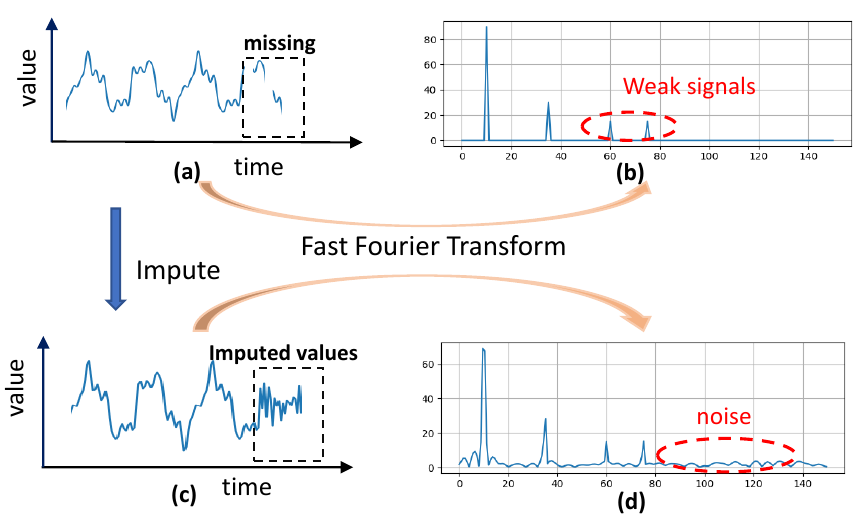}
\caption{(a) represents a time series with missing values. (b) represents a spectrogram. Noise is introduced into the raw spatial time series while imputing as shown in (c) compare to (a). (d) is a spectrogram associated with (c).
}
% Figure1(c)&(d) are the  backward process of diffusion model.In  figure1(d),the backward process is presented as polyline which is different from the curves in figure1(c) due to the finite diffusion time steps.Thus the final imputed series belongs to a different distribution from figure(c).
\label{fig1}
\end{figure}
Recently, deep learning methods are also widely adopted for spatiotemporal data imputation such as CNN and RNN based models \cite{wang2020deep}.The RNN based model BRITS \cite{BRITS} considers forward and backward connections between time steps in spatial time series imputation. However, biases can accumulate within the recursive structure of RNNs, ultimately impacting the model accuracy. ASTCMCN \cite{lijiyue} integrates Transformers with RNNs to model spatial correlations. CNN based models or GNN based models \cite{chen2020label} try to capture the spatial correlations of the time series by designing novel convolutional kernels.  TCN \cite{tcn} is a type of causal convolution, which convolves past states into the current state.

With the great success of attention mechanism in many areas, Transformer based models such as SAITS \cite{saits} and SPIN \cite{spin} implicitly capture the global temporal dependencies among nodes using cross-attention. However, Transformer generally models the spatial and temporal correlations of the spatial time series data separately, and thus may lose the time sequential properties. Recently, a Fourier transform-based deep method called TimesNet was proposed to adapt data shape based on the spectral characteristics to convolves temporal feature arbitrarily. However, it is still limited to the local receptive field of convolution kernels. The first diffusion model based method for spatial time series imputation is CSDI \cite{csdi}, and the following work SSSD \cite{sssd} utilizes the diffwave architecture \cite{diffwave} to effectively generate sequences. CSDI employs feature and temporal attention, while SSSD incorporates S4 \cite{s4} as a feature extractor. However, it is limited to its backbone. %Numerous approaches coexist, they all face some challenges.

As illustrated in Figure 1(a), spatial time series typically consist of a few signals which is related to spatial temporal relations. These signals represented in the frequency domain through fast fourier transform as several peaks in Figure 1(b). However, when encoding these sequences in the presence of missing data, noise is introduced by the encoder to be learned in model training, leading to the low-amplitude noise frequencies in the frequency domain plot and weakening the strength of the original signal. This implies that originally subtle signals become blurred. It is a common issue of deep models. But for diffusion model, the issue would be specially severe as diffusion model would randomly add noise in different intensity during forward process. Specially, as the forward process of diffusion model introduces noise step by step, the noise in each step has a different intensity from others. Diffusion model assumes all the noise follows the same Gaussian distribution, but in reality these generated noise may have a different mean value and intensity. This inconsistency leads to bias, causing the final data to deviate from the real data. This issue arises during the denoising process, rise the challenge of searching for non-major correlations (subtle signals), which is called noise imbalance. Although the diffusion model has shown promising results, they are still limited to the issue of noise imbalance.
% In summary, this involves two key issues. Firstly,spatial time series exhibit complex spatial-temporal dependencies.It lead to disabilities as conventional works mostly model spatial-temporal corrrelations separately and implicitly. 

To address the above limitations, we propose a learning framework named SaSDim coupled with Self-Adaptive Noise Scaling Diffusion Model. SaSDim integrates a novel High-Order Stochastic Differential Equations (SDEs) with an Across Spatio-Temporal Global Convolution (ASTGConv). This framework aims to adaptively balance noise intensity and model global spatiotemporal dependencies.
Specifically, SaSDim comprises three key modules: the Conditional Mixture Module, the Across Spatio-Temporal Global Convolution Module (ASTGConv), and the Probabilistic High-Order SDE(Stochastic Differential Equation) Solver Module. The Conditional Mixture Module aims to encode the spatial time series data to guide imputation. The ASTGConv module explicitly models the dependency of each node at different times. After the first two modules, we use the third module to balance the noise in each step. It is the Probabilistic High-Order SDE Solver Module , which generates and optimizes coefficient of high-order SDEs to change the training loss function, thus achieving noise scaling for balancing intensity during training. In addition, this coefficient is a learning parameter.
Inspired by SGConv \cite{SGConv}, the idea of ASTGConv involves explicitly representing the convolutional kernel as a superposition of waves (signals) that can present the frequency information of the time series. This enables direct modeling of semantic correlations between time steps. In our model, ASTGConv kernel is represented as waves infused with spatial information. These waves are combined through the action of the dynamic spatial kernel, reflecting dynamic interactions between locations (nodes). The primary contributions of this work are summarized as follows.

\begin{itemize}
\item
 we for the first time introduce a scaling mechanism to balance the noise for searching the non-major signals. Based on that, we propose a diffusion model based on  high-order SDEs solver, which both have the rigor on mathematics and heuristics in practice. %At the same time, it achieves a higher accuracy on spatial time series imputation.
 
 \item
The spatiotemporal correlations are modeled explicitly. Specially, the proposed across global spatio-temporal convolution AGSTConv can jointly and globally approximate the complex correlation between different time steps and different locations with explicit modeling.

\item
Our experiments on spatial time series imputation task demonstrate that the proposed probabilistic high-order SDEs solver module can effectively enhance conventional diffusion model denoisying and AGSTConv module can jointly capture the spatiotemporal correlations.

\end{itemize}

% The remainder of the paper is organized as follows. We will first briefly review related work in Section 2. Then, notations and problem definition will be introduced in Section 3. Next, we will introduce the proposed model in detail in Section 4, followed by experimental results in Section 5. Finally, we will conclude the paper in Section 6.

\section{RELATED WORK}
Early spatial time series imputation methods mostly relied on statistical techniques, such as K-Nearest Neighbors (KNN), Matrix Factorization (MF) \cite{MF}, and Multiple Imputation using Chained Equations. Deep learning methods have also been widely used in this domain. Deep autoregressive models based on recurrent neural networks (RNNs) are among the most popular approaches \cite{RNN1,RNN2,RNN3}. BRITS \cite{BRITS} is a representative model that utilizes bidirectional RNNs for imputation and using a simple linear regression layer to incorporate spatial information \cite{dlinear}. 

Modeling the global temporal dependency is challenging for RNN models as it only considers the local sequential relations. To address this issue, SAITS \cite{saits} introduced self-attention to capture the global temporal relations. SPIN further adopted a joint attention that combined spatial and temporal attentions to model information exchange between nodes. More recently, ASTCMCN, a deep model based on RNN, effectively combined transformer with RNN to capture both temporal and spatial dependencies. However, transformer-based models ignore the sequential relation of series.

Recently, motivated by the great success of generative models, researchers tried to impute missing data using generative-based model. GAIN \cite{gain} is a generative adversarial networks (GANs) for data imputation. Other methods like GAINFilling \cite{gainfilling} also rely on GANs to generate imputed sequences by matching the underlying data distribution. CSDI stands as a paradigmatic example of applying Denoising Diffusion Probabilistic Models (DDPM) \cite{ddpm} to data imputation. It sequentially employs feature and temporal attention to learn the noise at each step, while introducing conditions in the denoising process. However, this methods still suffer from the issue of parameter explosion as the increase of length of time series, making researchers to search lightweight and effective feature extractors for time series modeling.

Recently, State Space Models (SSM) \cite{s4d, s4nd, shashimi} have shown promising results in sequence modeling. The state space model (SSM) uses a set of linear dynamics equations to model nonlinear and physical systems with input, output, and state variables. Recently, a deep SSM-based model is proposed and outperformed several models. However, it suffers from a complex computation with the growth of the input sequences' lengths. Then a improved followup model is proposed to decomposes the state transition matrix
into the sum of low-rank and normal matrices and implements SSM as a global convolutional model. But these models are lack of heuristics due to its long mathematical proof. Inspired by these methods, \cite{SGConv} proposed SGConv, which generated global convolution kernels by upsampling local convolution kernels before concatenating them. SGConv effectively models the distance-dependent decays among nodes in a sequence and keeps the parameter scale sublinear with respect to the sequence length (naturally include time series). While achieving promising results, SGConv can not well catch the space information among graph nodes. This leads to limited performance in time series imputation.
More recently, a Fourier transform-based deep method called TimesNet has emerged. It adapts data shape adaptively based on the spectral characteristics of spatial time series to capture correlations within and across periods through multi-scale convolutions. However, it is still with the issue that the kernel only considers local information.

\section{Preliminaries}

{\bf Spatial graph }

Using the spatial distribution (latitude and longitude) of nodes, we calculate the spatial distance between each pair of nodes to construct a spatial graph. The adjacent matrix of the graph is then normalized into a Gaussian kernel. we call it the spatial graph, which is denoted as $G = (V, E)$, where $V = {v_1, ..., v_n }$ is a set of $N$ sensor
nodes deployed in the space and $E$ is a set of edges connecting these nodes. 
\newline
{\bf Spatial Time Series }

We denote $X \in R^{N \times D}$ as observations on spatial graph $G$. We denote the incomplete and complete spatial time series at time $t$, $X_{t} \in R^{N \times D} $and $Z_{t} \in R^{N \times D} $, respectively, 
where $N$ is the number of spatial nodes (e.g.  traffic sensors deployed at different locations) and $D$ is dimension of feature. 

The historical spatial time series can be represented as a sequence $X = 
(X_{t-k} , ..., X_{t})$. To conduct 
the spatial time series imputation over the locations where the data are unavailable,
we also define the mask matrix of incomplete data as follows:
\newline
\[ 
M_t (n) = 
\left\{
    \begin{array}{ll}
        0,~~if~~corresponding~~value~~is~~missing
        \\
        1, ~~otherwise
    \end{array}
\right.
\]

where $n$ denotes the sensor node, $M_t(n)$ is a mask function.
We mark the nodes without observations as 0 and the nodes with data as 1.

% \begin{definition}
% \textbf{}
% \end{definition}
{\bf Convolution theorem }
To use Fast Fourier Transform, we need introduce the convolution theorem, which is a communication theory as follows,
\begin{equation}
F(K \ast X) = F(K) \cdot F(X)
\end{equation}
where $\ast$ is a convolution operation, $\cdot$ is a multiply operation. $F$ is Fast Fourier Transform, which can be used to convolve time series.
\newline
{\bf Problem Definition }

Given the spatial graph $G = (V, E)$ and the corresponding incomplete spatial time series $X = (X_{0}, \ldots, X_{k})$, we aim to build a model $\epsilon_\theta$ to impute $X$ and obtain a complete spatial time series $Z = (Z_{0}, \ldots, Z_{k})$.

% \begin{equation}
% dx = -\frac{1}{2}\beta(t)(x_t - \nabla_x \log q_t(x_t))dt + \sqrt{\beta(t)}dw(t) \tag{1}
% \end{equation}

% \newline
% \begin{definition}
% \textbf{Complete spatial-temporal data}
% \end{definition}
% We denote $X \in R^{N × D}$ as urban flow observations on urban flow
% graph$ G$.We denote incomplete and complete urban flows at time $ 
% as t,X_{u,t} \in R^{N × D} $and $X_{c,t} \in R^{N × D} $respectively, 
% where $N$ is the number of nodes and $D$ is dimension. The 
% historical urban flows can be represented as a sequence $X_u = 
% (X_{u,t-k} , ..., X_{u,t})$ on the urban flow network. To conduct 
% the urban flow data imputation over the regions where the
% data are unavailable, we also define the mask matrix of incomplete 
% urban flows as follows:

%         $M_t (n) \begin{cases}
%             0, & $if X_{u,t} $is missing, \\
%             1, & $otherwise.$
%         \end{cases}
% (5)$
% \newline
% where $n$ denotes the sensor node,$M_t(n)$ is a mask function, which
% marks the nodes without urban flow observations as 0 and the nodes
% with data as 1.
% The complex spatial-temporal patterns are captured from the
% historical urban flow observations first and then help impute the
% missing data. However, some new spatial-temporal patterns that
% exist outside the urban flow graph G cannot be captured from the
% historical data. Thus, we design a super node outside the graph to
% represent new spatial-temporal data.

% \begin{equation}
% dx = -\frac{1}{2}\beta(t)(x_t - \nabla_x \log q_t(x_t))dt + \sqrt{\beta(t)}dw(t) \tag{1}
% \end{equation}

\section{Methodology}

The model framework of SaSDim is shown in Figure 2, which contains three modules, the Input \& Condition-mixture Module,
the Across Spatial-Temporal Global Convolution
Module and the Probabilistic High-Order SDE Solver module. Next, we will introduce the model in detail. %Totolly, we briefly name these first two module as incorporate computing block.

\subsection{Conditional Mixture Module}

In this module, we use a Conv1D encoder to embed sequence relation and spatial information to guide the generation of diffusion model. Specifically, the input data $X_t$ in Figure 2 which contains noise during denoising process, is first concatenated with a spatial time series data embedding that incorporates local information, which serves as a guidance for generation. Next, following the convolution theorem presented in preliminary, the concatenated tensor is computed to extract the frequency feature through fast fourier transform. And then it multiplied with convolution kernels from ASTGConv module and calculated from frequence feather to series. Additionally, diffusion embedding, feature embedding and time step embedding are included as supplementary information and added to the tensor after this operation.

Then, the tensor goes through a layer of gated activation units before entering the next layer of residual units. The results of several residual layers are added to the output, which maintains its original state through skip connections, and then fed into the probabilistic high-order SDE (stochastic differential equation) solver.
Briefly, we learn a function $\textbf{f}$ in each residual layer. $\textbf{f}$ gets the input $X_{L-1}$, and eventually outputs $X_L$.
\begin{equation}
X_L = \textbf{f}(X_{L-1})
\label{eq15} 
\end{equation}
where $L$ is the number of layers.

\subsection{Across Spatial-Temporal Global Convolution Module}

\begin{figure*}[!h]
\centering
\includegraphics[width=\linewidth]{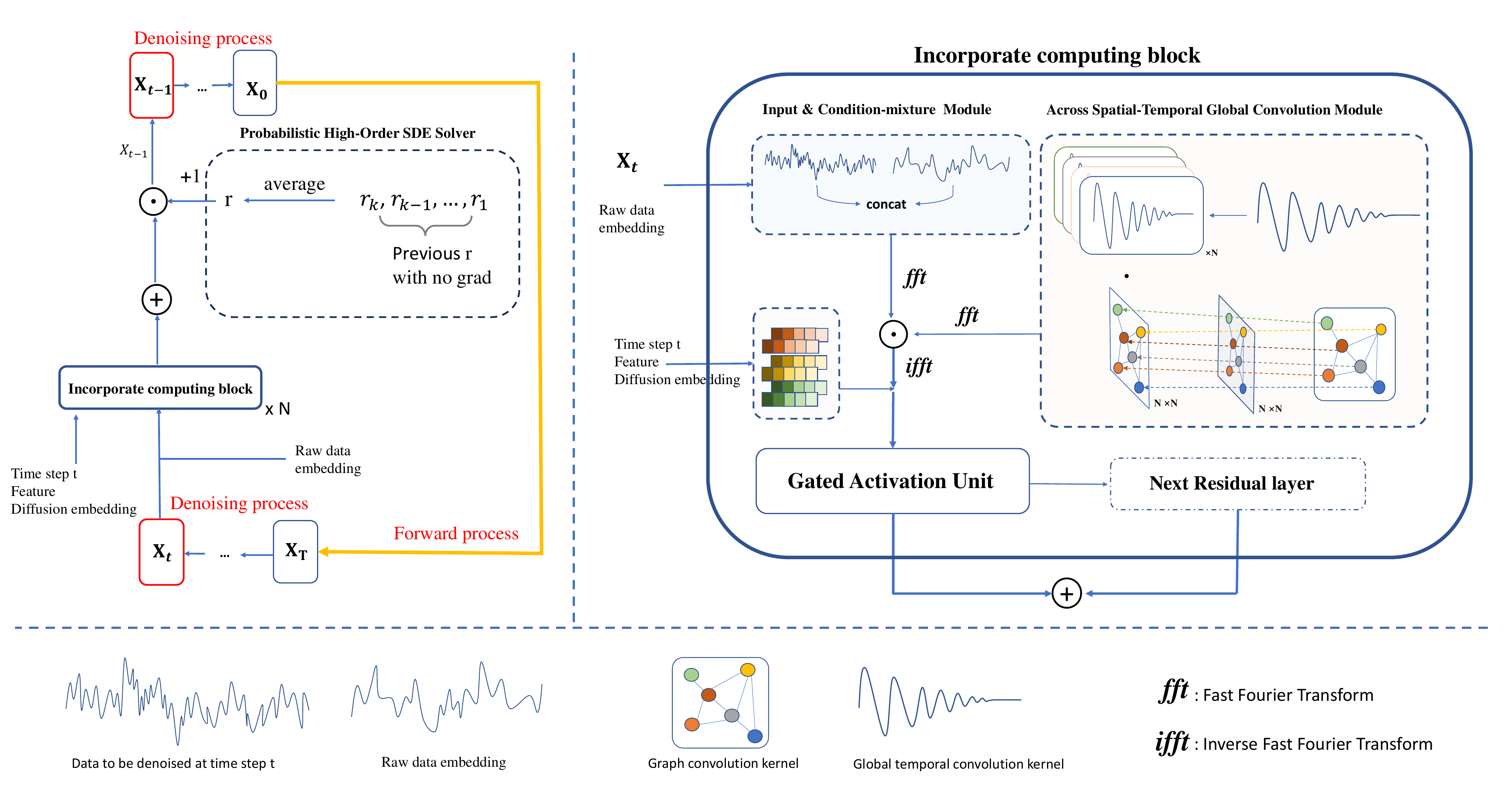}
\caption{The model framework of SaSDim. The right side of the figure shows the details of the Conditional Mixture Module and the Across Spatial-Temporal Global Convolution Module. The left side of the figure contains a module named Probabilistic High-Order SDE Solver Module. The bottom of the figure is the explanation of symbols.}
\label{fig1}
\end{figure*}

% Probabilistic High-Order SDE Solver Module.  Across Spatial-Temporal Global Convolution Module model adopt global temporal convolution kernel with spatial info to straightly model the correlation.Then we put the data from Incorporate computing block into Probabilistic High-Order SDE Solver Module so that we finally scale the predicted noise.} 

The Fast Fourier Transform (FFT) is an effective tool to extract frequency features from sequences to better show the correlations. In numerical computing, it is known as Discrete Fourier Transform (DFT) or FFT algorithm. DFT simplifies the numerous frequencies in a sequence to a smaller set of time series length $L$ components. This helps us calculate their weighted sum, similar to convolution.

Based on this idea and inspired by SGConv, we propose Across Spatial-temporal Global Convolution. Convolutional kernels gather information from nearby neighbors in a sequence, considering the ``closer means stronger'' rule \cite{SGConv}. For example, spatial graph convolutions group locations based on distances, and SGConv uses wave attenuation for temporal kernels. We further combine temporal kernels using graph spectrum information to build the across spatial-temporal convolutional kernels. These represent weighted sum of frequency components across nodes, which can explicitly model the spatiotemporal correlations.

Following this idea, we propose ASTGConv which includes the Global Temporal Convolution and Dynamic Graph Convolution.
First, the Global Temporal Convolution for modeling temporal correlation, generates convolutional kernels $K_{i}$ that build upon the wave curves with decay. Then we sum all kernels as follows,
\begin{equation}
Sum(K) = \frac{1}{Z} \cdot \sum_{i=0}^{N} K_i  \label{eq12} 
\end{equation}
% $\psi$ is the exponent kernel function,and $f$ is the normal function,$t$ is learnable parrameter,$A_{i}$ is a vector or a matrix.For the Exponential function of real or complex numbers, it can be defined as a Power series:$$e^x=1+\sum_{n=1}^{\infty}\frac{x^n}{n!}=1+x+\frac{x^2}{2!}+\frac{x^3}{3!}+\frac{x^4}{4!}+...$$ 。
This kernels gather information from nearby elements in time series. Following  the DFT algorithm, it convolves temporal info by weighted sum the frequencies.

We next introduce the Dynamic Graph Convolution for modeling the spatial correlations, which leverages spectral graph theory to transform the graph structure by altering the eigenvalues and trace of the Laplacian matrix. $A = P \lambda_A P^T $, where $P$ and $P^T$ are the eigenmatrix of matrix $A$ and $\lambda_A$ is the eigenvalues matrix. 
Then we have 

\begin{equation}
\bar{\lambda_A}= \lambda_A*W_{\lambda}\label{eq14}
\end{equation}

where $W_{\lambda}$ is a scaling factor which can be written as 

\begin{equation}
\quad W_{\lambda} = Diag(1) \odot (\alpha_0, \dots, \alpha_N) \label{eq14}
\end{equation}
\newline
where $(\alpha_0, \dots, \alpha_N)$ is the scaling factor corresponding to each eigenvalue. This transformation modifies the spatial correlation among them so that enables the ability for dynamically exchanging information between nodes.
Moreover, it scales the norm of the Laplacian matrix with $\alpha$, influencing the total sum of values within a specific set of nodes at a particular moment, expanding or contracting their influence.

Finally, we combine the Global Temporal Convolution and Dynamic Graph Convolution by element-wise multiplication to obtain the Cross Spatio-Temporal Global Convolution. This combined operation $\varphi_{\bar{A}}(K)$ integrates the strengths of both convolutions, allowing for more comprehensive spatio-temporal information extraction.
\begin{equation}
\varphi_{\bar{A}}(K) = \bar{A} K \label{eq14} 
\end{equation}
% According to Convolution Theorem, we can present theorem equations as:
In addition, $\bar{A}$ is calculated by $P \bar{\lambda_A} P^T$. According to Convolution Theorem and Eq(6), the theorem equation can be rewritten as
\begin{equation}
F(\varphi_{\bar{A}}(K) \ast X) = F(\varphi_{\bar{A}}(K)) \cdot F(X).
\end{equation}
Based on this theorem, we compute a mapping from time field to spectral field, and make them multiplied. Then we map the result back to time field. During the process, frequency components are weighted summed across nodes, where the kernel is regarded as the weight.

\subsection{Probabilistic High-Order SDE Solver
}
In this module, we first introduce the traditional stochastic differential equations (SDEs) that present the forward process. Then, we present our method that finds a new training loss to scale the imbalance noise to a proper level.   

First, SDEs perturb data to noise with a diffusion process governed by the following stochastic differential equation (SDE) \cite{diffusion}:
\begin{equation}
dx = f(x, t) dt + g(t) dw
\end{equation}
where $f(x, t)$ and $g(t)$ are diffusion and drift functions of the SDE, and $w$ is a standard Wiener process.
The forward process of DDPM (denoising diffusion probabilistic model) discretizes the SDE so that we could consider it as the limit of the following discrete form as $\Delta t \rightarrow{0} $:

\begin{equation}
\begin{aligned}
    x_{t+\Delta t} - x_t &= f_t(x_t)\Delta t + g_t \Delta t \sqrt{\epsilon} \\
    \epsilon &\sim N(0, I).
\end{aligned}
\end{equation}

% \begin{equation}
% dx = -\frac{1}{2}\beta(t)x dt + \sqrt{\beta(t)}dw

% \end{equation}
% And the generation process of DDPM \cite{ddpm} can be viewed as a particular consistency of the reverse-time DDPM.
For the sampling process, given $t^{'} = t + \Delta t$, then its discrete form can be written as:
\begin{eqnarray}
\begin{aligned}
\Delta x = -\left[ f_{t}(x_{t}) - g_{t}^2 \nabla x_{t} \log q(x_{t}) \right] \Delta t + g_{t} \sqrt{\Delta t }\epsilon 
\end{aligned}
\end{eqnarray}
Continualizing this form, we can get:
\begin{equation}
dx = [f(x, t) - g(t)^2 \nabla_x \log q_t(x)] dt + g(t) dw
\end{equation}
\newline
To achieve better results, we have computed the reverse process that includes higher-order derivatives.
\newline

\textit{\textbf{fact}:the training loss function can be represented as:$$
{\small \mathbb{E}_{Cond} \| \mathbf{s_{\theta}(x_t, t) - (1+r)\nabla_{x_t} \log q_{t}(x_t | x_0)} \|^2} 
$$}

To prove that, we need two lemmas written blow:

\textit{\textbf{lemma1}
High-order stochastic differential equation of equation(4) can be represented as:$$dx = \psi(\nabla_x)dt + g(t) dw -\frac{1}{2} * \beta_x * g(t)^2 \nabla_{xx} \log q_t(x_t)  dt$$
when $\Delta t \rightarrow 0$ and
$\psi(\nabla_x) = f(x, t) - g(t)^2 \nabla_x \log q_t(x)$. And $\beta_x$ is an upper bound of $\Delta_x$. }

More details see Appendix.

% \textit{Proof}:
% \newline
% Initially,the discretize form of high-order stochastic equation can be written as:
% $$\Delta x = \psi(\nabla_x)\Delta_t + g(t) dw -(\frac{1}{2!} * \Delta x * g(t)^2 \nabla_{xx} \log q_t(x_t) + \dots )\Delta_t$$ 
% to go futher,let's consider the third order item:
% $$\Delta x = \psi(\nabla_{xx})\Delta t + g(t) \Delta w -(\frac{1}{3!} (\Delta x)^2 g(t)^2 \nabla_{xxx} \log q_t(x_t) + \dots )\Delta t$$ 
% % $$p+q=c \Rightarrow ap+aq=ac \Rightarrow a(p+q)=ac \Rightarrow a^2+ab=ac$$
% % $$b^2+ab=bc$$
% % $$a^2+b^2+ab+ab=ac+bc \Rightarrow a^2+b^2=c(a+b)$$
% % 由于$a+b=c$，所以：
% % $$a^2+b^2=c^2$$
% % \noindent\textbf{Denoising Diffusion Probabilistic Models}
% % \cite{} 
% For the terms involving third-order derivatives, we have \(\Delta x\) raised to a higher power, and \(\Delta x\) is proportional to \(\sqrt{t}\). Therefore, as \(\Delta t\) approaches zero, the third-order terms become higher-order infinitesimals with respect to \(\Delta t\).
% By scaling, we can obtain:
% \begin{equation}
% \begin{split}
% dx = [f(x, t) - g(t)^2 \nabla_x \log q_t(x)] dt + g(t) dw
% \\ 
%  -\frac{1}{2} * \beta_x * g(t)^2 \nabla_{xx} \log q_t(x_t)dt \tag{7}
% \end{split}
% \end{equation}
% \hfill $\square$
% \newline
% \newline

Based on lemma1, we prove lemma2:
\newline

\textit{\textbf{lemma2}:Sampling equations of lemma1 can be represented as:$$\small \Delta x = -\left[ f_{t} - (1+r)g_{t}^2 \nabla x_{t} \log q(x_{t}) \right] \Delta t + g_{t} \sqrt{\Delta t}\epsilon $$as the traditional sampling equations is presented as below:$$\Delta x = -\left[ f_{t}(x_{t}) - g_{t}^2 \nabla x_{t} \log q(x_{t}) \right] \Delta t + g_{t} \sqrt{\Delta t }\epsilon.$$}

More details see Appendix.

Regarding the range of $r$ according to lemma2, we adopt a probabilistic sampling approach by sampling several points from a Gaussian distribution. Among these points, one is selected as the initial value for $r$. Considering that the magnitude of higher-order derivatives is smaller than that of the first-order derivative, the values of $r$ are confined within a window on the Gaussian distribution.

Subsequently, we update $r$ using the backpropagated gradients to make it approaching the true value gradually. Finally, taking into account that the value of $r$ computed by the BP algorithm is the mean of multiple training, we obtain the final sampled $r$ by averaging the $r$ values computed from each training iteration.
\subsection{Implementation Details}
The initialization of coefficient $r$ of high-order SDE is in the range from 0 to 0.2. We set the maximum noise level
to 0.02 and the layer of block to 6. The model is implemented using Pytorch and trained in an end-to-end manner using Adam with a learning
rate 0.001.

\section{Experiment}
\begin{table*}[h]
\renewcommand{\arraystretch}{1.3}
% 行间距放大为原来的1.5倍
\small
% \usepackage{multirow}
% \usepackage[normalem]{ulem}

% \usepackage{multirow}
% \usepackage[normalem]{ulem}

% \usepackage{multirow}
% \usepackage[normalem]{ulem}
% \usepackage{multirow}
% \usepackage[normalem]{ulem}
\begin{tabular}{cll|cccc|cccc|cccc}
\hline

\multicolumn{3}{l|}{\multirow{3}{*}{Model}} & \multicolumn{4}{c|}{AQI}                                                                                               & \multicolumn{4}{c|}{Nanjingyby}                                                                                       & \multicolumn{4}{c}{metr\_la}                                                                  \\ \cline{4-15} 
\multicolumn{3}{c|}{}                       & \multicolumn{2}{c|}{25\%}                 & \multicolumn{2}{c|}{50\%}      & \multicolumn{2}{c|}{25\%}                & \multicolumn{2}{c|}{50\%}     & \multicolumn{2}{c|}{25\%}                & \multicolumn{2}{c}{50\%}      \\ \cline{4-15} 
\multicolumn{3}{c|}{}                       & RMSE           & \multicolumn{1}{c|}{MAE} & RMSE           & MAE           & RMSE          & \multicolumn{1}{c|}{MAE} & RMSE          & MAE           & RMSE          & \multicolumn{1}{c|}{MAE} & RMSE          & MAE           \\ \hline
\multicolumn{3}{c|}{Transformer}            & 29.46          & 16.26                    & 31.49          & 17.45         & 3.91          & 1.71                     & 3.94          & 1.79          & 11.60         & 6.15                     & 12.13         & 6.39          \\
\multicolumn{3}{c|}{BRITS}                  & 28.76          & 15.72                    & 29.12          & 16.01         & 3.45          & 1.66                     & 3.68          & 1.75          & 10.12         & 5.59                     & 10.65         & 5.82          \\
\multicolumn{3}{c|}{SAITS}                  & 29.85          & 16.24                    & 30.97          & 17.18         & 3.69          & 1.50                     & 3.70          & 1.62          & 10.30         & 5.46                     & 10.51          & 5.73          \\
\multicolumn{3}{c|}{CSDI}                   & 14.52          & {\ul 7.71}               & 16.93          & {\ul 8.87}    & 3.86          & 1.68                     & 3.93          & 1.74          & 11.49         & 5.97                     & 12.08         & 6.32          \\
\multicolumn{3}{c|}{TimesNet}               & 21.01          & 12.38                    & 23.49          & 13.22         & {\ul 2.57}    & 1.42                     & \textbf{2.73} & 1.68          & 10.27         & 5.43                     & 10.64         & 5.69          \\
\multicolumn{3}{c|}{SPIN}                   & {\ul 12.98}    & 7.56                     & {\ul 16.53}    & 9.11          & 3.12          & {\ul 1.41}               & 3.34          & {\ul 1.58}    & {\ul 9.36}    & \textbf{4.92}            & {\ul 9.72}    & \textbf{5.08} \\
\multicolumn{3}{c|}{\textbf{SaSDim(ours)}}   & \textbf{11.21} & \textbf{7.03}            & \textbf{13.33} & \textbf{8.22} & \textbf{2.53} & \textbf{1.37}            & {\ul 2.96}    & \textbf{1.56} & \textbf{9.23} & {\ul 5.19}               & \textbf{9.60} & {\ul 5.36}    
\\ 
\hline
\end{tabular}
\caption{Comparison results between baselines and our method on AQI,METR-LA and Nanjingyby datasets. We report two error metrics
RMSE and MAE for two missing scenarios(25\%, 50\%).}
\end{table*}
\subsection{Datasets} 
We use the following three datasets for evaluation.

\textbf{METR-LA} is a dataset used in traffic flow prediction and imputation. It contains 207 traffic sensor nodes in Los Angeles County Highway with minute-level sampling rate.

\textbf{AQI} is collected from 36 AQI sensors distributed across the city of Beijing. This dataset serves as a widely recognized benchmark for imputation techniques and includes a mask used for evaluation that simulates the distribution of actual missing data \cite{mask}. For a specific month, such as January, this mask replicates the patterns of missing values from the preceding month. Across all scenarios, the valid observations that have been masked out are employed as targets for evaluation. We derive a spatial matrix from the spatial distribution for further analysis.

\textbf{Nanjingyby} is a visitor trip trajectory data collected from Nanjing Garden Expo Park. We artificially and randomly remove some values to mimic the missing values such as random missing. For this scenario, we set spatial time series values as 0 for the randomly selected regions or time slots. The details of the datasets are introduced as follows. Nanjingyby contains over 5 million crowd trip records in Nanjing Garden Expo Park from April 20 to June 30 in 2021. Each crowd trip includes id, regionid, time, latitude and longitude. 

In total, each dataset will be artificially masked 25\% or 50\% values in random. For the two datasets METR-LA and Nanjingyby, we partition the entire data into training, validation and testing sets by a ratio of 7 : 1 : 2.

\subsection{Evaluation Metrics}
We use two metrics Mean Absolute Error (MAE) and Root Mean Square Error (RMSE) defined as follows to evaluate the model performance.
\[
MAE = \frac{1}{T} \sum_{k=0}^{L} \left\| Z_{k} - Y_{k} \right\| 
\]
\[
RMSE = \sqrt{\frac{1}{T} \sum_{k=0}^{L} ( \left\| Z_{k} - Y_{k} \right\|^2_{F} }
\]
where $Z_{t}$ is the imputed value at time $t$ and $Y_{t}$ is the corresponding ground truth.

\subsection{Baselines} We compare our model with the following baselines.

\begin{itemize}

\item \textbf{SAITS} \cite{saits} is a method based on diagonally-masked self-attention (DMSA) and joint optimization.

\item \textbf{BRITS} \cite{BRITS} utilizes bidirectional RNN and MLP to integrate spatiotemporal information.

\item \textbf{TimesNet} \cite{timesnet} is a self-organized convolution model for time series data imputation task.

\item \textbf{SPIN} \cite{spin} employs threshold graph attention and temporal attention jointly to implicitly model the spatiotemporal sequences.

\item \textbf{CSDI} \cite{csdi} introduces the fractional diffusion model into the task of time series imputation, incorporating feature-mixing and channel-mixing transformers.

\end{itemize}

In addition, we add transformer \cite{att} as a baseline.
To study whether each module of SaSDim is useful, we also
compare SaSDim with the following variants for ablation study.

\begin{itemize}
\item SaSDim-lowO: This variant of SaSDim removes the probabilistic high-order SDE-solver module.

\item SaSDim-noC: SaSDim-noC  aggregates spatiotemporal correlations among nodes without the numeral constraint.
\item SaSDim-GConv: This variant only captures the temporal global relations to verify ASTGC module can capture spatiotemporal correlations.

\end{itemize}

% \begin{figure}[hbtp]

%   \centering
%   \includegraphics[width=0.7\linewidth]{figures/小图2.pdf}
%   \caption{ Frequency-domain representation of the imputed data(blue 
%  line) and Ground truth(grey line) }
%   \label{fig3}
% \end{figure}

\subsection{Experiment Result}

The results shown in Table 1 indicate that BRITS performs worst among all the baseline models because the MLP model cannot effectively capture the spatial correlations and the higher-order relationships between nodes without a good spatial encoding. SAITS shows slightly better performance, indicating that Diagonally-Masked Self-Attention (DMSA) can alleviate the impact of self-redundant information and improve the temporal prediction performance. SPIN outperforms other baseline models, suggesting that the improved attention mechanism with Hybrid Spatiotemporal Attention is effective in enhancing the performance of the original Transformer for spatial time series. The result of TimesNet results is slightly lower than those of SPIN from 1\% to 10\%, suggesting that while models that adaptively organize data for temporal convolution can effectively capture the temporal correlations, this convolution approach remains local and limits the ability of learning the global spatiotemporal dependencies. SaSDim achieves the best results in terms of two metrics in most cases with only three exceptions that achieve the second best results. Specifically, with a missing rate of 50\% on the AQI dataset, the MAE of SaSDim drops by 11\% compared to SPIN. This confirms that the ASTGConv module can better model the global spatial-temporal interactions between locations. SaSDim achieves the performance improvement by 7.5\% compared to CSDI. One can also see that CSDI performs as well as SaSDim on the AQI dataset but is surpassed by SaSDim by 20\% on the METR-LA dataset. This demonstrates the superiority of SaSDim in modeling spatial time series compared with existing diffusion model based approaches.
% \begin{figure}[hbtp]
%   \centering
%   \includegraphics[width=\linewidth]{figures/频域图.pdf}
%   \caption{frequency-domain representation of the imputed data(blue 
%  line) and Ground truth(grey line) }
%   \label{fig5
%   }
% \end{figure}
\subsection{Ablation Study}

\begin{figure}[!tp]
  \centering
  \includegraphics[width=0.7\linewidth]{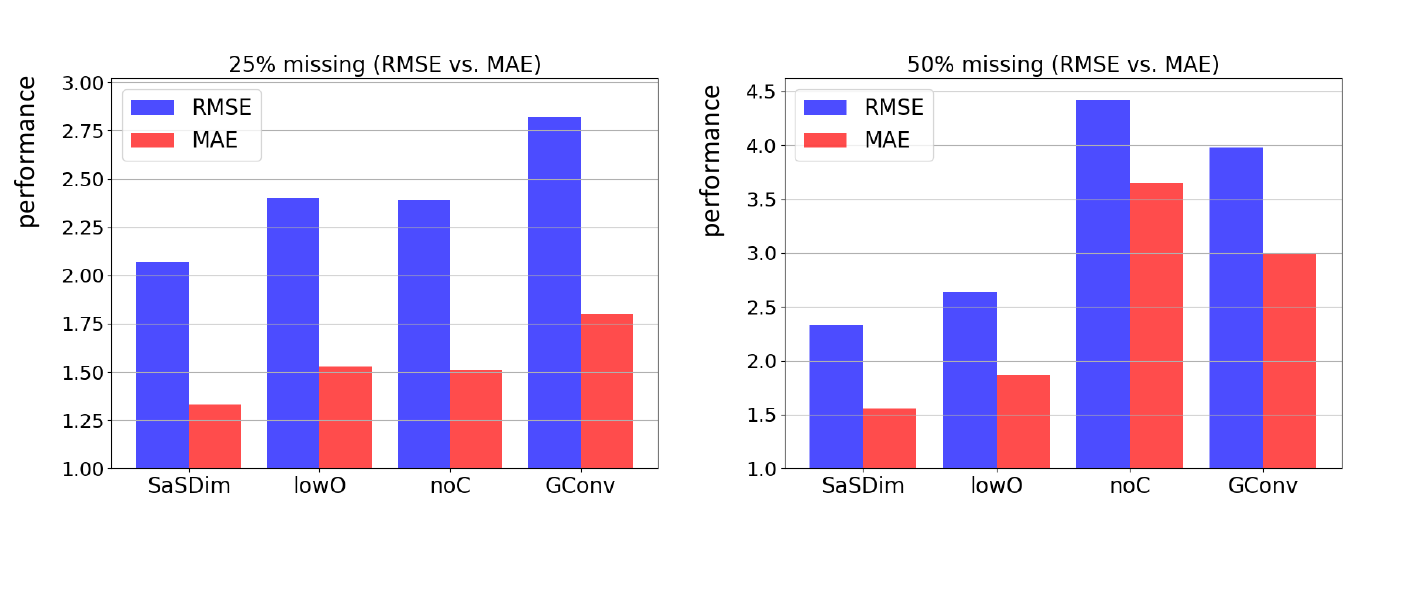}
  \caption{Evaluation among three variants on RMSE and MAE ( Nanjingyby).}
  \label{fig3}
\end{figure}

Figure 3 shows the distribution of RMSE and MAE bars for the three variants compared to SaSDim. It is evident that SaSDim-GConv performs significantly worse than the other variants by 25\% on 25\% data missing scenario, because it does not consider the geographical correlations between locations. The Constraint module is used to modulate the adjacency relationships between nodes. SaSDim-noConst(noC), which does not use spatial constraint conditions, achieves 12\% higher RMSE error than SaSDim on 25\% missing scenario and 90\% higher RMSE error on 50\% missing scenario. It confirms that spatial constraint conditions are helpful in aggregating the neighboring information. SaSDim-lowO directly applies first-order derivatives in the denoising process. Although SaSDim-lowO is superior to other variants, it still performs worse than SaSDim, significantly lagging behind SaSDim with a 15\% higher RMSE and 20\% higher MAE in the 25\% data missing scenario. This indicates that SaSDim can learn better representations through scaling down noise for denoising. This result verifies that the Across Spatial-temporal Global Convolution module and the Probabilistic High-Order SDE Solver module are both indispensable for improving the model performance.

% \begin{figure*}[h]
% \centering
% \includegraphics[width=\linewidth]{figures/模型大图.pdf}
% \caption{The framework of HoDiC, which containsConditional Mixture Module,Across Spatial-Temporal Global Convolution Module,and Probabilistic High-Order SDE Solver Module. }
% \label{fig1}
% \end{figure*}

\subsection{Parameter Study}
To analyze the parameter used in SaSDim, the scaling factor of the graph is initialized using a standard normal distribution. When the maximum value in the scaling factor array is larger than one, the values in the scaling factor array that are relatively smaller than 
the maximum value will become larger. Since the scaling factor array still follows a normal distribution, the difference between the maximum and minimum values will increase under equal probability. This leads to an increase in the variance of the convolutional kernels and an increase in uncertainty. When the minimum value in the scaling factor array is less than zero, it greatly increases the uncertainty of the graph. For an extreme example, when all the coefficients in the scaling factor array are negative, the adjacency matrix becomes a negative matrix, causing the matrix to lose its physical meaning.
% \begin{figure}[htbp]
%   \centering
%   \includegraphics[width=0.48\textwidth]{figures/小图3.pdf}
%   \caption{这是一个示例图片。}
%   \label{fig:example}
% \end{figure}

Next, we analyze the noise intensity which controls how much noise we would add. SaSDim uses a noise intensity of 0.5\cite{csdi} when imputing the air quality dataset. However, using the same noise intensity for the METR-LA and Nanjingyby datasets results in undesirable results. On the contrary, setting the intensity to 0.02 yields better results. This indicates that traffic flow data contains many subtle interaction signals, and high-intensity noise will completely mask these signals, making SaSDim difficult to impute them accurately.
\begin{figure}[!t]
  \centering
\includegraphics[width=0.7\linewidth]{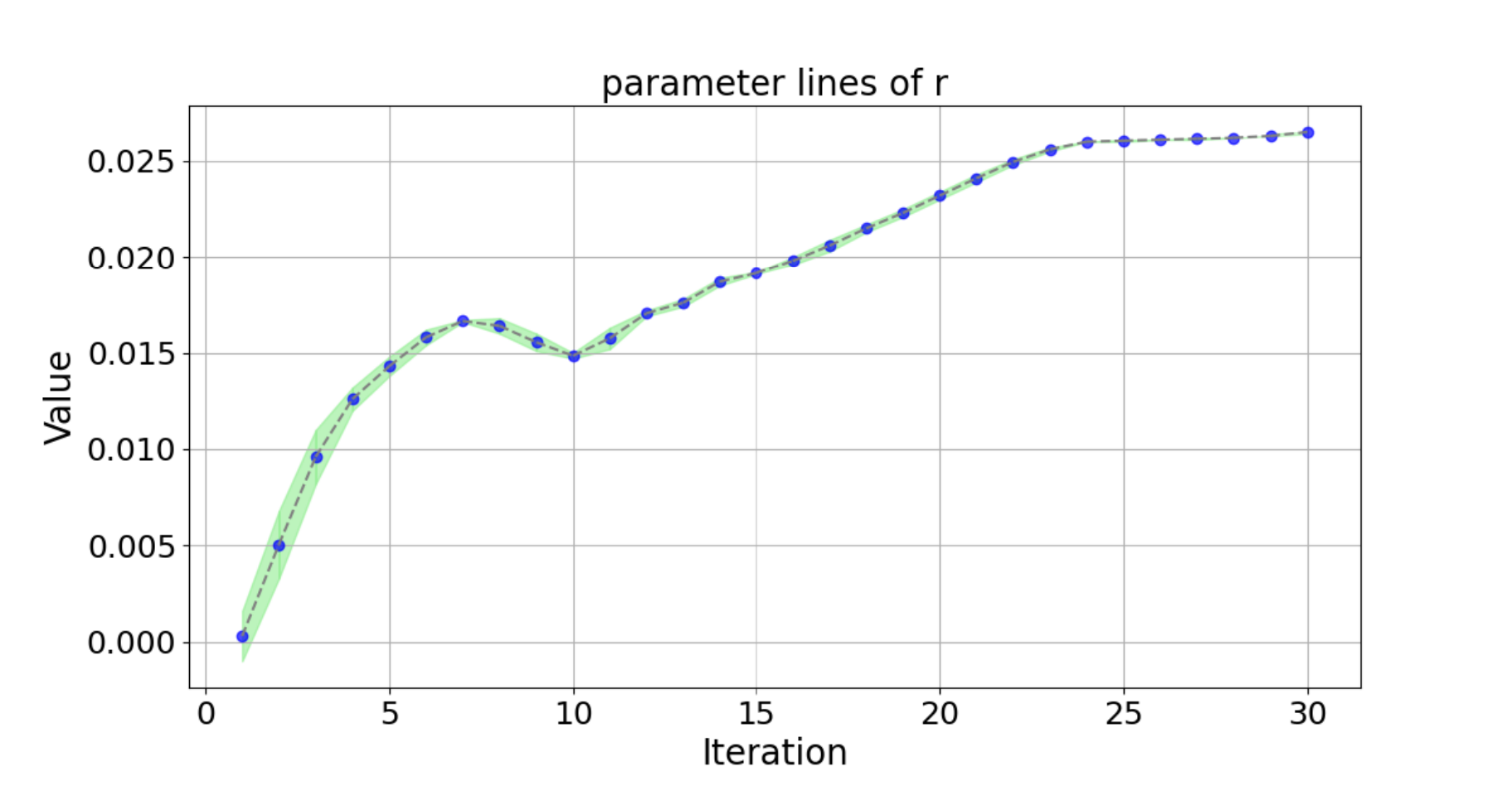}
  \caption{The orbit of coefficient $r$ appeared in the loss function. The blue point is a median value of all batch at an epoch. The green area is the range of variance.}
  \label{fig3}
\end{figure}

Finally, we study the effect of coefficients ($r$) of the probabilistic high-order SDE solver on the mode performance. According to the equation in Appendix, the absolute value of $r$ should be smaller than one, so we initially set the values of $r$ to a value in the range between -1 and 1. To better observe the trajectory and trend of $r$ during training, we set the initial value of $r$ to zero. The results show that $r$ gradually increases after a sudden decrease and eventually stabilizes around 0.026 as shown in Figure 4. This indicates that the noise scales down step by step. Next, to determine the final shared value of $r$ for sampling, we conducted two experiments. In one experiment, we set $r$ to be the last value obtained during optimization, and in the other experiment, we took the average of all the optimized $r$ values. The results show that setting $r$ to the last value obtained leads to significant fluctuations in performance, while using the average value yields stable results. This suggests that taking the average value mitigates the impact of randomness.
\begin{figure}[!t]
  \centering
  \includegraphics[width=0.7\linewidth]{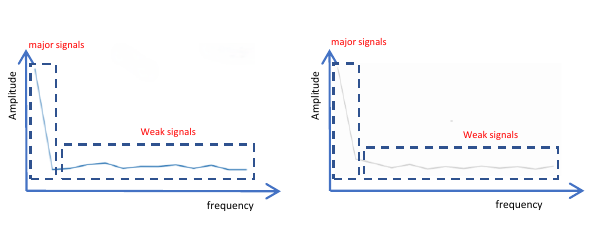}
  \caption{Visualization of frequency feather of both ground-truth (left) and imputed data (right) in Nanjingyby dataset. We randomly mask 25\% values.}
  \label{fig3}
\end{figure}

\subsection{Case Study}
We visualize the imputation results of one case in Figure 5 to further show the effectiveness of SaSDim. It shows that SaSDim captures the non-major signals as the polyline of weak signals fits the left one. Specially, We select a spatial time series from a specific location and transform it into the frequency domain using FFT. The high-amplitude frequency represents the major-signals. The left line in Figure 5 shows that the raw data has a few major signals and the most of signals are weak and non-major, which is consistent with the observations on the right side in Figure 5. Overall, SaSDim not only achieves good performances but imputes well intuitively.

\section{Conclusion}

This paper proposes a method called SaSDim inspired by global convolution and Denoising Diffusion Probabilistic Model learning framework for spatial time series imputation. SaSDim employs ASTGConv to capture the global dynamic interactions among locations. ASTGConv explicitly models the spatiotemporal correlations by generating dynamic spatiotemporal convolution kernels. SaSDim optimizes the coefficients $r$ of the high-order SDE solver during the training process to facilitate the generation of shared coefficients that scaling unbalanced noise.  Experimental results conducted on three real datasets verify the effectiveness of SaSDim.

%Bibliography
\nocite{*}
\bibliographystyle{unsrt}
\bibliography{references}

\begin{thebibliography}{10}

\bibitem{spin}
I.~Marisca, A.~Cini, and C.~Alippi, ``Learning to reconstruct missing data from
  spatiotemporal graphs with sparse observations,'' {\em Advances in Neural
  Information Processing Systems}, vol.~35, pp.~32069--32082, 2022.

\bibitem{RNN1}
Z.~Che, S.~Purushotham, K.~Cho, D.~Sontag, and Y.~Liu, ``Recurrent neural
  networks for multivariate time series with missing values,'' {\em Scientific
  reports}, vol.~8, no.~1, p.~6085, 2018.

\bibitem{RNN2}
Q.~Wang, S.~Ren, Y.~Xia, and L.~Cao, ``Bicmts: Bidirectional coupled
  multivariate learning of irregular time series with missing values,'' in {\em
  Proceedings of the 30th ACM International Conference on Information \&
  Knowledge Management}, pp.~3493--3497, 2021.

\bibitem{RNN3}
J.~Yoon, W.~R. Zame, and M.~van~der Schaar, ``Multi-directional recurrent
  neural networks: A novel method for estimating missing data,'' in {\em
  Proceedings of Time series workshop in international conference on machine
  learning}, 2017.

\bibitem{BRITS}
W.~Cao, D.~Wang, J.~Li, H.~Zhou, L.~Li, and Y.~Li, ``Brits: Bidirectional
  recurrent imputation for time series,'' {\em Advances in neural information
  processing systems}, vol.~31, 2018.

\bibitem{MF}
S.~Lee and D.~B. Fambro, ``Application of subset autoregressive integrated
  moving average model for short-term freeway traffic volume forecasting,''
  {\em Transportation research record}, vol.~1678, no.~1, pp.~179--188, 1999.

\bibitem{saits}
W.~Du, D.~C{\^o}t{\'e}, and Y.~Liu, ``Saits: Self-attention-based imputation
  for time series,'' {\em Expert Systems with Applications}, vol.~219,
  p.~119619, 2023.

\bibitem{lijiyue}
S.~Wang, J.~Li, H.~Miao, J.~Zhang, J.~Zhu, and J.~Wang, ``Generative-free urban
  flow imputation,'' in {\em Proceedings of the 31st ACM International
  Conference on Information \& Knowledge Management}, pp.~2028--2037, 2022.

\bibitem{timesnet}
H.~Wu, T.~Hu, Y.~Liu, H.~Zhou, J.~Wang, and M.~Long, ``Timesnet: Temporal
  2d-variation modeling for general time series analysis,'' {\em arXiv preprint
  arXiv:2210.02186}, 2022.

\bibitem{tcn}
P.~Hewage, A.~Behera, M.~Trovati, E.~Pereira, M.~Ghahremani, F.~Palmieri, and
  Y.~Liu, ``Temporal convolutional neural (tcn) network for an effective
  weather forecasting using time-series data from the local weather station,''
  {\em Soft Computing}, vol.~24, pp.~16453--16482, 2020.

\bibitem{att}
A.~Vaswani, N.~Shazeer, N.~Parmar, J.~Uszkoreit, L.~Jones, A.~N. Gomez,
  {\L}.~Kaiser, and I.~Polosukhin, ``Attention is all you need,'' {\em Advances
  in neural information processing systems}, vol.~30, 2017.

\bibitem{sssd}
J.~M.~L. Alcaraz and N.~Strodthoff, ``Diffusion-based time series imputation
  and forecasting with structured state space models,'' {\em arXiv preprint
  arXiv:2208.09399}, 2022.

\bibitem{SGConv}
Y.~Li, T.~Cai, Y.~Zhang, D.~Chen, and D.~Dey, ``What makes convolutional models
  great on long sequence modeling?,'' {\em arXiv preprint arXiv:2210.09298},
  2022.

\bibitem{csdi}
Y.~Tashiro, J.~Song, Y.~Song, and S.~Ermon, ``Csdi: Conditional score-based
  diffusion models for probabilistic time series imputation,'' {\em Advances in
  Neural Information Processing Systems}, vol.~34, pp.~24804--24816, 2021.

\bibitem{s4}
A.~Gu, K.~Goel, and C.~R{\'e}, ``Efficiently modeling long sequences with
  structured state spaces,'' {\em arXiv preprint arXiv:2111.00396}, 2021.

\bibitem{s4d}
A.~Gu, K.~Goel, A.~Gupta, and C.~R{\'e}, ``On the parameterization and
  initialization of diagonal state space models,'' {\em Advances in Neural
  Information Processing Systems}, vol.~35, pp.~35971--35983, 2022.

\bibitem{s4nd}
E.~Nguyen, K.~Goel, A.~Gu, G.~Downs, P.~Shah, T.~Dao, S.~Baccus, and C.~R{\'e},
  ``S4nd: Modeling images and videos as multidimensional signals with state
  spaces,'' {\em Advances in neural information processing systems}, vol.~35,
  pp.~2846--2861, 2022.

\bibitem{shashimi}
K.~Goel, A.~Gu, C.~Donahue, and C.~R{\'e}, ``It’s raw! audio generation with
  state-space models,'' in {\em Proceedings of International Conference on
  Machine Learning}, pp.~7616--7633, PMLR, 2022.

\bibitem{dlinear}
A.~Zeng, M.~Chen, L.~Zhang, and Q.~Xu, ``Are transformers effective for time
  series forecasting?,'' in {\em Proceedings of the AAAI conference on
  artificial intelligence}, vol.~37, pp.~11121--11128, 2023.

\bibitem{diffwave}
Z.~Kong, W.~Ping, J.~Huang, K.~Zhao, and B.~Catanzaro, ``Diffwave: A versatile
  diffusion model for audio synthesis,'' {\em arXiv preprint arXiv:2009.09761},
  2020.

\bibitem{gain}
J.~Yoon, J.~Jordon, and M.~Schaar, ``Gain: Missing data imputation using
  generative adversarial nets,'' in {\em Proceedings of International
  conference on machine learning}, pp.~5689--5698, PMLR, 2018.

\bibitem{gainfilling}
Y.~Luo, X.~Cai, Y.~Zhang, J.~Xu, {\em et~al.}, ``Multivariate time series
  imputation with generative adversarial networks,'' {\em Advances in neural
  information processing systems}, vol.~31, 2018.

\bibitem{ddpm}
J.~Ho, A.~Jain, and P.~Abbeel, ``Denoising diffusion probabilistic models,''
  {\em Advances in neural information processing systems}, vol.~33,
  pp.~6840--6851, 2020.

\bibitem{diffusion}
L.~Yang, Z.~Zhang, Y.~Song, S.~Hong, R.~Xu, Y.~Zhao, Y.~Shao, W.~Zhang, B.~Cui,
  and M.-H. Yang, ``Diffusion models: A comprehensive survey of methods and
  applications,'' {\em arXiv preprint arXiv:2209.00796}, 2022.

\bibitem{mask}
X.~Yi, Y.~Zheng, J.~Zhang, and T.~Li, ``St-mvl: filling missing values in
  geo-sensory time series data,'' in {\em Proceedings of the 25th International
  Joint Conference on Artificial Intelligence}, 2016.

\bibitem{wang2020deep}
S.~Wang, J.~Cao, and S.~Y. Philip, ``Deep learning for spatio-temporal data
  mining: A survey,'' {\em IEEE transactions on knowledge and data
  engineering}, vol.~34, no.~8, pp.~3681--3700, 2020.

\bibitem{chen2020label}
H.~Chen, Y.~Xu, F.~Huang, Z.~Deng, W.~Huang, S.~Wang, P.~He, and Z.~Li,
  ``Label-aware graph convolutional networks,'' in {\em Proceedings of the 29th
  ACM international conference on information \& knowledge management},
  pp.~1977--1980, 2020.

\bibitem{wang2016estimating}
S.~Wang, L.~He, L.~Stenneth, S.~Y. Philip, Z.~Li, and Z.~Huang, ``Estimating
  urban traffic congestions with multi-sourced data,'' in {\em Proceedings of
  2016 17th IEEE International conference on mobile data management (MDM)},
  vol.~1, pp.~82--91, IEEE, 2016.

\end{thebibliography}

\appendix
\section{Appendix}

\textit{\textbf{lemma1}:High-order stochastic differential equation of equation(4) can be represented as:$$dx = \psi(\nabla_x)dt + g(t) dw -\frac{1}{2} * \beta_x * g(t)^2 \nabla_{xx} \log q_t(x_t)  dt$$
when $\Delta t \rightarrow 0$ and
$\psi(\nabla_x) = f(x, t) - g(t)^2 \nabla_x \log q_t(x)$. And $\beta_x$ is an upper bound of $\Delta_x$.}

\textit{Proof}:

Initially,the discretize form of high-order stochastic equation can be written as:
\begin{equation}
\begin{split}
\Delta x = \psi(\nabla_x)\Delta_t + g(t) dw - \\(\frac{1}{2!} * \Delta x * g(t)^2 \nabla_{xx} \log q_t(x_t) + \dots )\Delta_t
\end{split}
\end{equation}
to go futher,let's consider the third order item:
\begin{equation}
\begin{split}
\Delta x = \psi(\nabla_{xx})\Delta t + g(t) \Delta w - \\ (\frac{1}{3!} (\Delta x)^2 g(t)^2 \nabla_{xxx} \log q_t(x_t) + \dots )\Delta t
\end{split}
\end{equation}

For the terms involving third-order derivatives, we have \(\Delta x\) raised to a higher power, and \(\Delta x\) is proportional to \(\sqrt{t}\). Therefore, as \(\Delta t\) approaches zero, the third-order terms become higher-order infinitesimals with respect to \(\Delta t\).
By scaling, we can obtain:
\begin{equation}
\begin{split}
dx = [f(x, t) - g(t)^2 \nabla_x \log q_t(x)] dt + g(t) dw
\\ 
 -\frac{1}{2} * \beta_x * g(t)^2 \nabla_{xx} \log q_t(x_t)dt
\end{split}
\end{equation}
where $\beta_x$ is the upper bound of $\Delta x$.
\hfill $\square$
\newline
Based on lemma1, we prove lemma2:
\newline
\newline

\textit{\textbf{lemma2}:Sampling equations of lemma1 can be represented as:$$\small \Delta x = -\left[ f_{t} - (1+r)g_{t}^2 \nabla x_{t} \log q(x_{t}) \right] \Delta t + g_{t} \sqrt{\Delta t}\epsilon $$as the traditional sampling equations is presented as below:$$\Delta x = -\left[ f_{t}(x_{t}) - g_{t}^2 \nabla x_{t} \log q(x_{t}) \right] \Delta t + g_{t} \sqrt{\Delta t }\epsilon.$$}

\textit{Proof}:
\newline
Given that 
$q(x_t|x_0) = N(x_t; \bar{\alpha_{t}}  x_0, \bar{\beta_{t}}^2 I)$  ,where $\bar{\beta_{t}}$ is the total noise intensity coefficient and
then we can gain the equation below:

\begin{equation}
\nabla x_t \log q(x_t|x_0) = -\frac{x_t - \bar{\alpha_{t}}  x_0}{\bar{\beta_{t}}^2} = -\frac{\bar{\epsilon}}{\bar{\beta}_{t}}
 \tag{8}
\end{equation}
Then the higher-order derivatives can be represented as:
\begin{center}
\begin{equation}
  \nabla_{xx} \log q_t(x_t | x_0) * \Delta x_t = \alpha (\frac{\bar{\epsilon}}{\beta_{t+\Delta t}} - \frac{\bar{\epsilon}}{\beta_t})= r_1 * \bar{\epsilon} \tag{9}
\end{equation}
\end{center}
as $\Delta x_t \sim  O(\Delta(\sqrt{t})) $ ,$\alpha 
$  is limited to a finite real number.
according to lemma1:
% $$\nabla_{xx} \log q_t(x_t)$$
\begin{equation}
\begin{aligned}
    \Delta x = -\left[ f_{t}(x_{t}) - g_{t}^2 \nabla x_{t} \log q(x_{t}) \right] \Delta t+ 
    \\
    \beta_x * g(t)^2 \nabla_{xx} \log q_t(x_t) \Delta t
    + g_{t} \sqrt{\Delta t }\epsilon
\end{aligned}
\end{equation}
 and we gain $\nabla x_t \log q(x_t)$ as equation below:

\begin{equation}
\begin{split}
\nabla x_t \log q(x_t) =\frac{\mathbb{E}_{x_0}[\nabla x_t q(x_t|x_0)]}{\mathbb{E}_{x_0}[q(x_t|x_0)]}  
\\
= \frac{\mathbb{E}_{x_0}[q(x_t|x_0) \nabla x_t \log q(x_t|x_0)]}{\mathbb{E}_{x_0}[q(x_t|x_0)]} 
\end{split}
\end{equation}
then we gain the result only by computing $ q(x_t|x_0)$.Also, it's in the same way to compute $\nabla_{xx_t} \log q(x_t)$ with the rule of differentiation of fractions and the equation (7).

\begin{equation}
\begin{split}
\nabla_{xx} \log q(x_t) 
= \frac{\mathbb{E}_{x_0}[q(x_t|x_0) \nabla_{xx} \log q(x_t|x_0)]}{\mathbb{E}_{x_0}[q(x_t|x_0)]} 
\end{split}
\end{equation}

\begin{equation}
\begin{aligned}
 \nabla x_{t} \log q(x_{t})]+\beta_x * \nabla_{xx} \log q_t(x_t) =
 \\
\frac{\mathbb{E}_{x_0}[q(x_t|x_0)(\nabla_{x} \log q(x_t|x_0) - \beta_x \nabla_{xx} \log q(x_t|x_0))]}{\mathbb{E}_{x_0}[q(x_t|x_0)]}
\end{aligned}
\end{equation}

given $s_{\theta}(x_t, t) = \epsilon_{\theta}$,then the training loss can be presented as follows:

\begin{center}
\begin{equation}
 \mathbb{E}_{Cond} \| \mathbf{s_{\theta}(x_t, t) - (1+r)\nabla_{x_t} \log q_{t}(x_t | x_0)} \|^2
\end{equation}
\end{center}

% \begin{center}
% \begin{equation}
% {\small \mathbb{E}_{ x_0 \sim q(x_0), x_t \sim q(x_t | x_0)} \| \mathbf{s_{\theta}(x_t, t) - \nabla_{x_t} \log q_{t}(x_t | x_0)} \|^2} \tag{10}
% \end{equation}
% \end{center}

given $t^{'} = t + \Delta t$,then the sampling process can be written as:

$$\small \Delta x = -\left[ f_{t} - (1+r)g_{t}^2 \nabla x_{t} \log q(x_{t}) \right] \Delta t + g_{t} \sqrt{\Delta t}\epsilon $$
\hfill $\square$

 % file mwe.bib

\end{document}